\pdfoutput=1

\documentclass[11pt,letterpaper]{article}

\usepackage{EMNLP2022}

\usepackage{times}
\usepackage{latexsym}

\usepackage[T1]{fontenc}

\usepackage[utf8]{inputenc}

\usepackage{microtype}

\usepackage{inconsolata}

\usepackage{todonotes}
\usepackage{multirow}
\usepackage{amssymb,amsmath}
\usepackage{booktabs}
\usepackage{colortbl}
\usepackage{tablefootnote}
\DeclareUnicodeCharacter{0327}{\ensuremath{\forall}}

\title{JoeyS2T: Minimalistic Speech-to-Text Modeling with JoeyNMT}

\author{Mayumi Ohta \\
Computational Linguistics \\
Heidelberg University, Germany\\
  \texttt{ohta@cl.uni-heidelberg.de} \\\And
  Julia Kreutzer \\
  Google Research \\
  \texttt{jkreutzer@google.com} \\\And
  Stefan Riezler \\
  Computational Linguistics \& IWR \\
  Heidelberg University, Germany \\
  \texttt{riezler@cl.uni-heidelberg.de}}

\begin{document}
\maketitle
\begin{abstract}
 JoeyS2T is a JoeyNMT \citep{kreutzer-etal-2019-joey} extension for speech-to-text tasks such as automatic speech recognition and end-to-end speech translation.  It inherits the core philosophy of JoeyNMT, a minimalist NMT toolkit built on PyTorch, seeking simplicity and accessibility. JoeyS2T's workflow is self-contained, starting from data pre-processing, over model training and prediction to evaluation, and is seamlessly integrated into JoeyNMT's compact and simple code base. On top of JoeyNMT's state-of-the-art Transformer-based encoder-decoder architecture, JoeyS2T provides speech-oriented components such as convolutional layers, SpecAugment, CTC-loss, and WER evaluation. Despite its simplicity compared to prior implementations, JoeyS2T performs competitively on English speech recognition and English-to-German speech translation benchmarks. 
The implementation is accompanied by a walk-through tutorial and available on \url{https://github.com/may-/joeys2t}.
\end{abstract}

\section{Introduction}
End-to-end models recently have been shown to be able to outperform complex pipelines of individually trained components in many NLP tasks. For example, in the area of automatic speech recognition (ASR) and speech translation (ST), the performance gap between end-to-end models and cascaded pipelines, where an acoustic model is followed by an HMM for ASR, or an ASR model is followed by a machine translation (MT) model for ST, seems to be closed \citep{sperber-2019-attention, bentivogli-etal-2021-cascade}. An end-to-end approach has several advantages over a pipeline approach: First, it mitigates error propagation through the pipeline. Second, its data requirements are simpler since intermediate data interfaces to bridge components can be skipped. Furthermore, intermediate components such as phoneme dictionaries in ASR or transcriptions in ST need significant amounts of additional human expertise to build. For end-to-end models, the overall model architecture is simpler, consisting of a unified end-to-end neural network. Nonetheless, end-to-end components can be initialized from non end-to-end data, e.g., in audio encoding layers \citep{xu-etal-2021-stacked} or text decoding layers \citep{li-etal-2021-multilingual}.

ASR or ST tasks usually have a higher entry barrier than MT, especially for novices who have little experience in machine learning, but also for NLP researchers who have previously only worked on text and not speech processing. This can also be seen in the population of the different tracks of NLP conferences. For example, the ``Speech and Multimodality'' track of ACL 2022 had only a third of the number of papers in the ``Machine Translation and Multilinguality'' track.\footnote{\url{https://public.tableau.com/views/ACL2022map/Dashboard1?:showVizHome=no}} However, thanks to the end-to-end paradigm, those tasks are now more accessible for students or entry-level practitioners without huge resources, and without the experience of handling the different modules of a cascaded system or speech processing. The increased adoption of Transformer architectures \citep{vaswani17} in both text \citep{DBLP:journals/corr/abs-2108-05542} and speech processing \citep{speech-transformer,Karita2019ACS,Karita2019ImprovingTE} has further eased the transfer of knowledge between the two fields, in addition to making joint modeling easier and more unified. 

Reviewing existing code bases for end-to-end ASR and ST---for example, DeepSpeech \citep{hannun2014deepspeech}, ESPnet \citep{inaguma-etal-2020-espnet, watanabe2020espnet}, fairseq S2T \citep{wang-etal-2020-fairseq}, NeurST \citep{zhao-etal-2021-neurst} and SpeechBrain \citep{ravanelli2021speechbrain}---it becomes apparent that the practical use of open-source toolkits still requires significant experience in navigating large-scale code, using complex data formats, pre-processing, neural text modeling, and speech processing in general. High code complexity and a lack of documentation are frustrating hurdles for novices. We propose JoeyS2T, a minimalist and accessible framework, to help novices get started with speech recognition and translation, to accelerate their learning process, and to make ASR and ST more accessible and transparent, that is directly targeting novices and their needs.

We hope that making more accessible implementations will also have trickle-down effects of making the research built on top of it more accessible and more linguistically and geographically diverse \citep{joshi-etal-2020-state}. This effect has already been observed for the adoption of JoeyNMT for text MT for low-resource languages \citep{forall-nekoto-etal-2020-participatory,camgoz2020sign,zhao-etal-2020-automatic,zacarias-marquez-meza-ruiz-2021-ayuuk,DBLP:journals/corr/abs-2106-15115,mirzakhalov2021evaluating}. Furthermore, speech technology has an even higher potential for language inclusivity \citep{CMUWilderness,abraham-etal-2020-crowdsourcing,zhang-etal-2022-nlp,liu-etal-2022-always}.

\section{Speech-to-Text Modeling}\label{sec:task}

Automatic speech recognition and translation require mapping a speech feature sequence $X = \{\mathbf{x}_i \in \mathbb{R}^d\}$ to a text token sequence $Y = \{y_t \in \mathcal{V}\}$. The continuous speech signal in its raw wave form is pre-processed into a sequence of discrete frames that are each represented as $d$-dimensional speech feature vectors $\mathbf{x}_i$, e.g., log Mel filterbanks at the $i$-th time frame. In contrast, a textual sequence is naturally composed of discrete symbols that can be broken down into units of different granularity, e.g. characters, sub-words, or words. These units then form a vocabulary, so in the above formulation $y_t$ is the $t$-th target token from the vocabulary $\mathcal{V}$. The goal of S2T modeling is then to find the most probable target token sequence $\hat{Y}$ from all possible vocabulary combinations $\mathcal{V}*$:
\begin{align}\label{eq:s2t}
    \hat{Y} &= \underset{Y\in \mathcal{V}*}{\arg\max} \;p(Y\mid X).
\end{align}

\subsection{Why End-to-End Modeling?}
In conventional HMM modeling, the posterior probability $p(Y\mid X)$ from Eq.~\ref{eq:s2t} is decomposed into three components by introducing the HMM state sequences $S=\{s_t\}$:
\begin{align}
    p(Y\mid X) &\approx \underbrace{p(X\mid S)}_{\text{Acoustic Model}} \underbrace{p(S\mid Y)}_{\text{Lexical Model}} \underbrace{p(Y)}_{\text{LM}}.
\end{align}
The components correspond to an acoustic model $p(X\mid S)$, a lexical representation model $p(S\mid Y)$, and a language model $p(Y)$.

For practitioners, this means that three individual models need to be implemented, trained and combined. This comes with a large overhead, since each of them requires dedicated linguistic resources and experience in training and tuning. Attention-based deep neural networks have reduced this burden significantly since they implicitly model all three components in a single neural network, mapping $X$ directly to $Y$ \citep{chorowski-2015-attention, chan-2016-listen}.

\subsection{Optimization}
Most approaches to sequence-to-sequence learning tasks like MT use the cross-entropy (Xent) loss for optimization, and break the sequence prediction task down to a token-level objective. The posterior probability from above is modeled as the product of output token probabilities conditioned on the entire input sequence $X$ and the target prefix $y_{<t}$:
\begin{equation}
    p_{\mathrm{xent}}(Y\mid X) := \prod_{t} p(y_t\mid y_{<t}; X).
\end{equation}

A popular alternative in ASR is to employ Connectionist Temporal Classification (CTC) loss \citep{graves2014ctc}. CTC uses a Markov assumption to model the transition of states similar to conventional HMM:

\begin{equation}
    p_{\mathrm{ctc}}(Y\mid X) := \sum_{\mathcal{A}} \prod_{t} p(a_t\mid X),
\end{equation}
where $\mathcal{A}$ denotes the set of valid alignments from $X$ to $Y$, $a_t\in\mathcal{A}$ is one possible alignment at the $t$-th time step, and marginalizing the conditional probability $p(a_t\mid X)$ over all valid possible alignments yields the sequence-level probability.

This CTC formulation is suitable to learn monotonic alignments between audio and text, and it also can handle very long sequences efficiently by solving dynamic programming on the state transition graph. The assumption of conditional independence at different time steps is a potentially harmful simplification which is compensated for by a token-level objective and by jointly minimizing cross-entropy and CTC loss \citep{hori-etal-2017-joint,watanabe-2017-hybrid}. The final optimization objective in the JoeyS2T implementation is a logarithmic linear combination of the label-smoothed cross-entropy loss and the CTC loss defined above:
\begin{align}
    \mathcal{L}_{\mathrm{total}} := &(1-\lambda) \log p_{\textrm{xent}}(Y\mid X) \nonumber\\
    &\hspace{1em}+ \lambda \; \log p_{\textrm{ctc}}(Y\mid X),
\end{align}
where $\lambda \in [0,1]$ is an interpolation parameter.

\section{Design Principles}\label{sec:design}
\paragraph{Simplicity:} We devoted considerable effort to keep JoeyS2T's module structure simple and flat. It directly employs the PyTorch \citep{paszke2019pytorch} backend and has a low level of abstraction (details in Section~\ref{sec:complexity}). JoeyS2T has a minimal list of external dependencies that can be easily installed via the \texttt{PyPI}\footnote{\url{https://pypi.org/}} tool. Even for pre-processing, external dependencies on tools such as Kaldi \citep{povey2011kaldi} are avoided. For filterbank feature extraction, we use TorchAudio\footnote{\url{https://github.com/pytorch/audio}} which is seamlessly integrated into PyTorch. In contrast to other toolkits, speech modules extended in JoeyS2T are only built for speech-to-text modeling. It does not implement speech enhancement, nor speaker detection or speech generation. While this might appear like a limitation, we believe that the reduction of functionalities to a carefully identified minimum for ST and ASR is the key for increased accessibility.\footnote{A clean code base can always be extended by users once they are more proficient. For example, JoeyNMT has been successfully extended to other modalities and integrated into web interfaces by advanced users. See \url{https://github.com/joeynmt/joeynmt\#projects-and-extensions}}

\paragraph{Accessibility:} We also have written extensive documentation and walk-through tutorials to help newcomers become more familiar with speech technologies. JoeyS2T also provides pretrained models including configuration files which lower the barrier to get started.
To guarantee the accessibility of the code, we open-sourced JoeyS2T under a very permissive license (Apache 2.0). 
 The JoeyS2T developer community actively supports user questions and requests. We maintain an open platform to discuss bug fixes, possible extensions etc. All contributions are first automatically controlled by the internal unit tests and will manually be reviewed by our team.

\paragraph{Reproducibility:}
To ensure that the reported results are comparable and reproducible, we release models trained on publicly available data. Our evaluation metrics are described in detail (tokenization, punctuation handling etc.). All pre- and post-processing scripts are published with a data download path and explicit hyperparameter configurations. We track all code changes in our repository and provide version information which is often a critical factor for reproducibility as bug fixes can affect evaluation scores.

\section{Implementation and Usage}\label{sec: JoeyS2T}
\subsection{Hyperparameter Configuration} 
JoeyS2T sets up experiments based on a YAML-style configuration file which declares the whole pipeline, just like JoeyNMT. Processes are run in a Python interface without relying on external Bash or Perl scripts. 
In the configuration file, users can choose between the tasks \texttt{MT} (Machine Translation) or \texttt{S2T} (Speech-to-Text) in order to inform JoeyS2T about the input data type: audio or text. The hyperparameters of speech-related modules such as SpecAugment, 1d-Conv etc. can also be specified in the same configuration file.\footnote{Sample configuration files for different datasets are available at \url{https://github.com/may-/joeys2t/configs}}

\subsection{Data Loading and Pre-processing}
\paragraph{Source Audios:} 
We separated computationally heavy pre-processing steps from model training, e.g., the conversion from raw wave forms to spectrograms by Fourier transformation. We employ the TorchAudio API to extract audio features in the pre-processing scripts. JoeyS2T includes modules for Cepstral Mean Variance Normalization (CMVN) \citep{viikki1998cepstral} and SpecAugment \citep{park2019specaugment} by default. These are applied minibatch-wise before the input data are fed into the encoder.
 
\paragraph{Data Loading:} As a precautionary measure to avoid memory allocation errors (which can happen for large audio inputs) we implemented on-the-fly data loading: we only store the path to the data in the iterator, and load the actual spectrogram features into memory every time a minibatch is constructed.

\paragraph{Target Texts:} 
For target texts, we expect users to prepare a tokenization model independently and to specify the path to the trained tokenizer. Besides rule-based character-level tokenization and basic white space splitting, we currently support subword-nmt tokenizers \citep{sennrich-etal-2016-neural} and SentencePiece tokenizers \citep{kudo-richardson-2018-sentencepiece}. Users can specify tokenizer options in JoeyS2T's configuration file. 
During training, JoeyS2T applies text tokenization on the fly. Since the text length can be calculated only after tokenization, instance filtering by length is applied in this step. Thanks to this flexible on-the-fly tokenization, dynamic data augmentation methods i.e., BPE Dropout \citep{provilkov-etal-2020-bpe}, SwitchOut \citep{wang-etal-2018-switchout} or ADA \citep{lam-2021-ada} can be easily integrated.

\subsection{Architectures}

\begin{figure}
\centering
\includegraphics[width=\columnwidth,trim={0em 4.2em 4em 2em},clip]{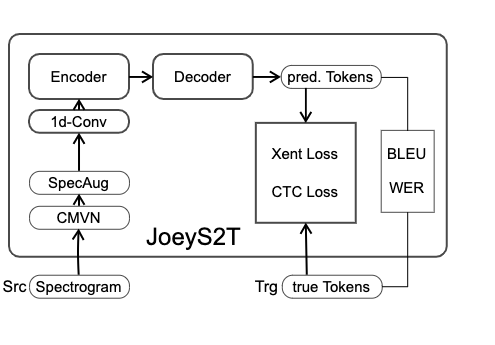}
\caption{Architecture of JoeyS2T. We reuse JoeyNMT's basic building blocks and extended them by essential audio-specific modules.}
\label{fig:architecture}
\end{figure}

JoeyS2T supports a Transformer-based encoder-decoder architecture (see Figure \ref{fig:architecture}). We reuse the self-attention encoder and decoder layers of JoeyNMT, and modify them in order to support speech-specific components.

\paragraph{Input Representations:}
Instead of converting token embeddings from discrete one-hot encodings to continuous vectors (as done for text input), we directly feed the sequence of filterbank vectors to the encoder. The embedding size in text-based JoeyNMT thus corresponds to the filterbank frequency size in JoeyS2T.

\paragraph{Encoder:} The biggest difference to the original text-to-text Transformer architecture is the 1-dimensional convolution layer (1d-Conv) placed before the self-attention encoder. It compresses potentially redundant features along the time dimension in order to capture phonetic structures. Each 1d-Conv layer has a stride of $2$. This further downsamples the sequence by a factor of $2^l$, where $l$ is the number of 1d-Conv layers. The reduction of the input length is essential for computation speed: Speech feature sequences are usually much longer than text token sequences, and the computational complexity of one self-attention block is $\mathcal{O}(u^2\cdot d)$~\cite{vaswani17}, where $u$ is the maximal input length (number of tokens in textual input, or number of time frames in speech input), and $d$ is the embedding size.

\paragraph{Decoder:}
We reuse the decoder construction of the original JoeyNMT code, but add one additional linear layer for the CTC loss on top of the self-attentive decoder layers.

\paragraph{Inference:}
We support greedy and beam search based on the token probability distributions. All inference enhancements introduced in JoeyNMT v2.0 such as repetition penalty, n-gram blocker, probability scoring, attention visualization of cross-attention heads in transformer layers, etc. are supported by JoeyS2T as well.

\begin{table*}[!ht]
\centering
\begin{tabular}{l|c|llll}
\hline
&& \multicolumn{4}{c}{\textbf{LibriSpeech 100h} (WER $\downarrow$)}\\
System & Architecture& \texttt{dev-clean} & \texttt{dev-other} & \texttt{test-clean}& \texttt{test-other} \\
\hline
\citet{kahn-2020-self}\textsuperscript{$\dagger$} &BiLSTM& 14.00& 37.02& 14.85 & 39.95\\
\citet{laptev-2020-you}\textsuperscript{$\dagger$}  &Transformer&10.3 & 24.0 & 11.2 & 24.9 \\
\href{https://github.com/espnet/espnet/tree/master/egs2/librispeech_100/asr1}{ESPnet}\textsuperscript{$\ddagger$} &Transformer& \hphantom{1}8.1& 20.2& \hphantom{1}8.4&  20.5 \\
\href{https://github.com/espnet/espnet/tree/master/egs2/librispeech_100/asr1}{ESPnet}\textsuperscript{$\ddagger$} &Conformer& \hphantom{1}6.3& 17.4& \hphantom{1}6.5&  17.3 \\
JoeyS2T &Transformer& 10.66 $\pm$ 0.36& 23.82 $\pm$ 0.34& 12.02 $\pm$ 0.32& 24.75 $\pm$ 0.37\\
\hline
\hline
&& \multicolumn{4}{c}{\textbf{LibriSpeech 960h} (WER $\downarrow$)}\\
System &Architecture& \texttt{dev-clean} & \texttt{dev-other} & \texttt{test-clean}& \texttt{test-other}\\
\hline
\citet{gulati-2020-conformer}\textsuperscript{$\dagger$} &Conformer& 1.9 & 4.4 & 2.1 & 4.9\\
\href{https://github.com/espnet/espnet/tree/master/egs2/librispeech/asr1#without-lm}{ESPnet}\textsuperscript{$\ddagger$} &Conformer& 2.3 & 6.1 & 2.6 & 6.0\\
\href{https://huggingface.co/speechbrain/asr-transformer-transformerlm-librispeech}{SpeechBrain}\textsuperscript{*} &Conformer& 2.13& 5.51& 2.31& 5.61\\
\href{https://huggingface.co/facebook/s2t-small-librispeech-asr}{fairseq S2T}\textsuperscript{*} &Transformer& 3.23 & 8.01 & 3.52& 7.83\\
\href{https://huggingface.co/facebook/wav2vec2-base-960h}{fairseq wav2vec2}\textsuperscript{*} &Conformer& 3.17& 8.86 &3.39 &8.57 \\
JoeyS2T &Transformer& 3.79 $\pm$ 0.27& 8.84 $\pm$ 0.39& 4.31 $\pm$ 0.52& 8.66 $\pm$ 0.35\\
\hline
\end{tabular}
\caption{
Averaged results in WER on the English \textbf{LibriSpeech} dataset over three runs with standard deviations ($\pm$). We compute the WER on lowercased transcriptions without punctuations using SacreBLEU's 13a tokenizer. $\dagger$: results were reported in the papers linked above. $\ddagger$: results were taken from the repository linked above. *: we downloaded their pretrained models from the repository, and ran the inference and the evaluation on the same test data as we use in JoeyS2T.}\label{tab:librispeech}
\end{table*}

\begin{table*}[!ht]
\centering
\begin{tabular}{l|cc|ll|ll}
\hline
& \multicolumn{2}{c|}{MuST-C ver.}& \multicolumn{2}{c|}{\textbf{ASR} (WER $\downarrow$)} & \multicolumn{2}{c}{\textbf{MT} (BLEU $\uparrow$)} \\
System &train&eval& \texttt{tst-COMMON} & \texttt{tst-HE} & \texttt{tst-COMMON} & \texttt{tst-HE}\\
\hline
\citet{Gangi2019}\textsuperscript{$\dagger$} & v1 & v1 &27.0 & - & 25.3& -\\
\citet{zhang-etal-2020-adaptive}\textsuperscript{$\dagger$} & v1 & v1 & - & - & 29.69 & - \\
\href{https://github.com/espnet/espnet/tree/master/egs/must_c}{ESPnet}\textsuperscript{$\ddagger$} & v1 & v1 &12.70 & - & 27.63 & - \\
\href{https://github.com/facebookresearch/fairseq/blob/main/examples/speech_to_text/docs/mustc_example.md}{fairseq S2T}\textsuperscript{*} & v1 & v1 &12.72 & 10.93 & - & - \\
 JoeyS2T & v2 & v1 &  18.86$\pm$0.37& 15.19$\pm$0.56& 23.07$\pm$0.14& 20.21$\pm$0.17\\
\hline
\href{https://github.com/facebookresearch/fairseq/blob/main/examples/speech_to_text/docs/mustc_example.md}{fairseq S2T}\textsuperscript{*} & v1& v2& 11.88& 10.43& -& -\\
 JoeyS2T & v2& v2& 12.95$\pm$0.32& 11.16$\pm$0.31& 27.17$\pm$0.63& 24.85$\pm$0.68\\
\hline\hline
& \multicolumn{2}{c|}{MuST-C ver.}& \multicolumn{2}{c|}{\textbf{Cascade ST} (BLEU $\uparrow$)} & \multicolumn{2}{c}{\textbf{End2End ST} (BLEU $\uparrow$)} \\
System &train&eval& \texttt{tst-COMMON} & \texttt{tst-HE} & \texttt{tst-COMMON} & \texttt{tst-HE}\\
\hline
\citet{Gangi2019}\textsuperscript{$\dagger$} & v1 & v1 & 18.5 & - & 17.3 & -\\
\citet{zhang-etal-2020-adaptive}\textsuperscript{$\dagger$} & v1 & v1 & 22.52 & - & 20.67 & - \\
\href{https://github.com/espnet/espnet/tree/master/egs/must_c}{ESPnet}\textsuperscript{$\ddagger$} & v1 & v1 & - & - & 22.91 & - \\
\href{https://github.com/facebookresearch/fairseq/blob/main/examples/speech_to_text/docs/mustc_example.md}{fairseq S2T}\textsuperscript{*} & v1 & v1 & - & - & 22.70 & 21.70 \\
JoeyS2T & v2 & v1 & 21.89$\pm$0.64& 21.03$\pm$0.66& 20.53$\pm$0.29& 21.13$\pm$0.46\\
\hline
\href{https://github.com/facebookresearch/fairseq/blob/main/examples/speech_to_text/docs/mustc_example.md}{fairseq S2T}\textsuperscript{*} & v1& v2& -& -& 23.20& 22.23\\
 JoeyS2T & v2& v2& 23.95$\pm$0.59& 22.65$\pm$0.58& 23.33$\pm$0.39& 22.90$\pm$0.69\\
\hline
\end{tabular}
\caption{\label{tab:mustc}
Averaged results on the \textbf{MuST-C en-de} dataset over three runs with standard deviations ($\pm$). We compute the BLEU on truecased translations with punctuations using SacreBLEU's 13a tokenizer.\footnotemark\, $\dagger$: results were reported in the papers linked above. $\ddagger$: results were taken from the repository linked above. *: we downloaded their pretrained models from the repository, and ran the inference and evaluation on the same test data as we use in JoeyS2T.}
\end{table*}

\footnotetext[6]{\texttt{nrefs:1|case:mixed|eff:no|tok:13a|smooth:exp|version:2.1.0}}

\subsection{Evaluation Metrics}
JoeyS2T supports Character F-score (ChrF)~\citep{popovic-2015-chrf}, BLEU~\citep{papineni-etal-2002-bleu} and Word Error Rate (WER) based on Levenshtein distance \citep{gonzalo-2001-levenshtein} as evaluation metrics for ASR and ST. We import  \texttt{sacrebleu}\footnote{\url{https://github.com/mjpost/sacrebleu}} \citep{post-2018-call} for ChrF and BLEU, and \texttt{editdistance}\footnote{\url{https://github.com/roy-ht/editdistance}} \citep{hyyro-2001-explaining} for WER. In addition, perplexity and accuracy can be monitored during training on Tensorboard \citep{tensorflow2015-whitepaper}.

\subsection{Documentation and Tutorial}
We follow the documentation strategy of JoeyNMT, which means that all extended functions have their own docstring and in-line comments for tensor shapes. Unit tests covering essential modules are automatically triggered on every commit to the repository.

In the hands-on tutorial, we present working examples for ASR and ST as Jupyter notebooks.\footnote{Demo video: \url{https://youtu.be/bpBtq2jLolQ}} The walk-through tutorial is self-contained and explains the whole pipeline: installation steps, data downloading, data pre-processing, configuration, model training/fine-tuning, inference and evaluation. We will keep the tutorial up to date with potential future API changes.

\begin{table}[ht!]
\begin{small}
\centering
\begin{tabular}{lrrr}
\toprule
 \multicolumn{2}{r}{\textbf{ESPnet2}\footnotemark} & \textbf{fairseq}\footnotemark & \textbf{JoeyS2T}\footnotemark\\
\midrule
Python files & 287 & 407 & 24 \\
Code lines& 41427 & 65097 & 5450\\
Comment lines & 10260 & 11042 & 2137\\
\midrule
Comment/Code Ratio &  0.25& 0.17&0.39 \\
\bottomrule
\end{tabular}
\caption{\label{complexity}
Code complexity measured using \url{https://github.com/AlDanial/cloc} v1.94.}
\end{small}
\end{table}
\footnotetext[10]{\url{https://github.com/espnet/espnet/tree/master/espnet2} (commit hash \texttt{039cc5d})}
\footnotetext[11]{\url{https://github.com/pytorch/fairseq/tree/main/fairseq} (commit hash \texttt{ad3bec5})}
\footnotetext[12]{\url{https://github.com/may-/joeys2t/tree/main/joeynmt} (commit hash \texttt{a80802a})}
 
\subsection{Code complexity}\label{sec:complexity}
JoeyNMT exhibits the spirit of minimalism by aiming to achieve 80\% of the output quality with 20\% of a common toolkit’s code size (80/20 principle; \cite{pareto1896cours}). Table \ref{complexity} gives statistics on code complexity. In terms of the numbers of Python files and code lines, JoeyS2T is 10--11 times more compact than ESPnet \citep{inaguma-etal-2020-espnet, watanabe2020espnet} and fairseq \citep{wang-etal-2020-fairseq}. 
However, both ESPnet and fairseq are general-purpose toolkits, covering a wide range of tasks beyond MT, ASR or ST, such as language modeling or speech synthesis, while JoeyS2T is designed for a speech-to-text tasks only. Yet JoeyS2T's comment-to-code ratio is much higher than that of the competitors.

JoeyS2T offers a flat code structure in order to make debugging along the stack trace easier and to reduce the number of code files and nested classes/functions to read through. In contrast, fairseq's codebase is organized hierarchically. This deep hierarchy comes from the structured class inheritance, which is an important component of object-oriented programming for experienced developers. However, such hierarchical class inheritance is sometimes a big stumbling block for novices \citep{wiedenbeck1999comparison}. We intentionally abandon deeply inherited class design and use novice-friendly flat structure instead. As a result, developers do not have to allocate their cognitive resources to framework-specific software design principles, but they can concentrate on the logic they want to realize. JoeyS2T encourages novices to dive into speech-to-text research before they mature in high-context system design such as hierarchical class inheritance or decorators.

\section{Experimental Results on Benchmarks}\label{sec:benchmarks}
Despite its simplicity, JoeyS2T achieves a performance on standard benchmarks that is comparable to other high-functional speech-to-text toolkits.

\subsection{ASR on LibriSpeech}

LibriSpeech \citep{panayotov-2015-ribrispeech} is the de-facto standard English ASR benchmark that contains 960 hours 
of audiobooks in Project Gutenberg. The corpus is publicly available under the CC BY 4.0 license and many works set their goal to achieve state-of-the-art WER on its test splits.

Tables \ref{tab:librispeech} present the results of models trained on 100h and 960h audio, respectively. JoeyS2T shows comparable performance with current Transformer-based models, which are generally outperformed by Conformer \cite{gulati-2020-conformer} models. 

\subsection{ST on MuST-C}

MuST-C \citep{cattoni2021must} is a publicly available speech translation corpus built from English TED Talks. It consists of English transcriptions and translations into 14 languages, contributed by volunteers. We trained our model on the English-German subset of version 2, and evaluated the model both on version 1 and version 2 \texttt{tst-COMMON}, and \texttt{tst-HE} splits.

MuST-C is a challenging dataset due to its spontaneous speech that contains hesitations, disfluent utterances, etc. on the source side. Furthermore, the ground-truth target texts derived from the subtitles are also noisy. There are some additional descriptions of non-verbal information, i.e., ``(applause)'' ``(laughter)'', or ``♪ (music)''. Those are not actually pronounced in the source, but provided in the target, which makes learning more difficult. We normalized such noisy expressions and specified them as special tokens during the subword training, so that they are not tokenized into subwords but kept as single tokens. For the sake of reproducibility, we provide a preprocessing script for all normalization steps.

For ST tasks, we first pretrained ASR models and MT models using the gold transcriptions. Then we initialized the encoder layers of an end-to-end ST model with the pretrained ASR encoders and the decoder layers with the pretrained MT decoders, and further trained it on the end-to-end ST task.

The ST results can be found in Table \ref{tab:mustc}. JoeyS2T shows competitive results, both in end-to-end scenarios and in a cascade using the same pre-trained models. We also include the ASR and MT pretraining results for reference.

\section{Conclusion \& Future Work}
We described JoeyS2T, an extension of the JoeyNMT toolkit to the spoken language processing tasks ASR and ST. JoeyS2T is characterized by its minimalist design, prioritization of simplicity, accessibility and reproducibility in its code and documentation. The code is self-contained and requires minimal prior experience with speech or language processing. In benchmark evaluations, JoeyS2T performed comparable or superior to other ASR or ST code bases, while having much lower code complexity.

While its functionality is kept minimal, support for state-of-the-art architectures such as wav2vec and Conformer might be desired for future extensions. 

\section*{Limitations}
The limitations of our work mainly concern the reproducibility of comparable state-of-the-art results. First, there are many different preprocessing variants which are quite complex (length filtering, speed shift, lowercasing, punctuation normalization etc.) and not always clearly documented. Second, the same problem appears in evaluation. There is no commonly accepted evaluation scheme (including lower-cased vs. true-cased results, with or without punctuation, etc.). While the \texttt{sacrebleu} library is a first step to addressing this problem in MT, we believe that the speech processing community also needs such efforts to standardize speech-to-text evaluation.

Since the goal of our work is not to present a new state-of-the-art in speech-to-text modeling, we did not invest a large effort into hyperparameter tuning, but only varied three different random seeds in our setup, and used the default settings for competitor systems.

\section*{Acknowledgements}
We would like to thank the members of the StatNLP group at Heidelberg University and the AIMS Senegal students for their feedback on the tutorial. Furthermore, we appreciate the discussions with the Masakhane\footnote{\url{https://www.masakhane.io/}} community in the early stages of the toolkit development. We also thank Yaraku Inc.\footnote{\url{https://www.yarakuzen.com/}} for the opportunity to publish JoeyS2T tutorial articles.\footnote{\url{https://atmarkit.itmedia.co.jp/ait/articles/2208/17/news002.html}}

\bibliography{anthology,emnlp2022}
\bibliographystyle{acl_natbib}
\end{document}